\documentclass[preprint,3p]{elsarticle}

\usepackage{booktabs}

\usepackage{amssymb,amsthm,amsmath}
\usepackage{subcaption}
\usepackage{graphicx}
\usepackage{subfig}
\usepackage{float}
\usepackage{multirow}
\usepackage{longtable,lscape}
\usepackage{cases}
\usepackage{url}
\usepackage{caption}
\usepackage{color}

\usepackage{amsthm}
\usepackage[pagewise]{lineno}
\usepackage{xcolor}

\newcount\Comments  % 1 suppresses notes to selves in text
\definecolor{darkgreen}{rgb}{0,0.5,0}
\newcommand{\kibitz}[2]{\ifnum\Comments=0\textcolor{#1}{#2}\fi}

\usepackage{fancyhdr}
\usepackage{nomencl}
\makenomenclature

\usepackage{etoolbox}
\renewcommand\nomgroup[1]{%
 \item[\bfseries
  \ifstrequal{#1}{A}{States and Actions}{%
  \ifstrequal{#1}{B}{Car-following Models}{%
  \ifstrequal{#1}{C}{Other Notations}{}}}%
]}

\usepackage{array}
\usepackage{multirow}
\usepackage{multicol}
\usepackage{makecell}
\usepackage{comment}
\usepackage{graphicx}
\usepackage{soul}

\usepackage{booktabs,caption}
\usepackage[flushleft]{threeparttable}
\usepackage{arydshln}
\usepackage{wrapfig}

\usepackage[ruled,vlined,linesnumbered]{algorithm2e}

\usepackage{enumerate}

\usepackage{makecell}%To keep spacing of text in tables
\setcellgapes{4pt}%parameter for the spacing
\usepackage[export]{adjustbox}
\usepackage{fullpage}

\theoremstyle {plain}% default

\newcounter{x}

\theoremstyle{definition}

\theoremstyle{remark}

\usepackage{titlesec}
\titleformat{\paragraph}
{\normalfont\normalsize\bfseries}{\theparagraph}{1em}{}
\titlespacing*{\paragraph}
{0pt}{3.25ex plus 1ex minus .2ex}{1.5ex plus .2ex}

\usepackage{color}

\makeatletter
\def\ps@pprintTitle{%
	\let\@oddhead\@empty
	\let\@evenhead\@empty
	\def\@oddfoot{\reset@font\hfil\thepage\hfil}
	\let\@evenfoot\@oddfoot
}
\makeatother

\numberwithin{equation}{section}

\journal{Transportation Research Part C Special Issue ``Managing Future Motorway and Urban Traffic Systems"}

\begin{document}

\begin{frontmatter} 
		
		
		
\title{Discovering the Precursors of Traffic Breakdowns Using Spatiotemporal Graph Attribution Networks}

\date{\today}
		
		
\author[cu]{Zhaobin Mo}

\author[min]{Xiangyi Liao}

\author[anl]{Dominik A. Karbowski}

\author[asu]{Yanbing Wang\corref{cor}}\ead{yanbing.wang@asu.edu}

\cortext[cor]{Corresponding author.}

\address[cu]{Department of Civil Engineering and Engineering Mechanics, Columbia University}
\address[min]{School of Economics, Minzu University of China}
\address[anl]{Vehicle and Mobility Systems Department, Argonne National Laboratory}
\address[asu]{School of Sustainable Engineering and the Built Environment, Arizona State University}

\begin{abstract}

Understanding and predicting the precursors of traffic breakdowns is critical for improving road safety and traffic flow management. This paper presents a novel approach combining spatiotemporal graph neural networks (ST-GNNs) with Shapley values to identify and interpret traffic breakdown precursors. By extending Shapley explanation methods to a spatiotemporal setting, our proposed method bridges the gap between black-box neural network predictions and interpretable causes. We demonstrate the method on the Interstate-24 data, and identify that road topology and abrupt braking are major factors that lead to traffic breakdowns.

\end{abstract}
		
\end{frontmatter}

\section{Introduction}

\begin{wrapfigure}{r}{0.4\textwidth}
\vspace{-0.9cm}
  \begin{center}      \includegraphics[width=0.38\textwidth]{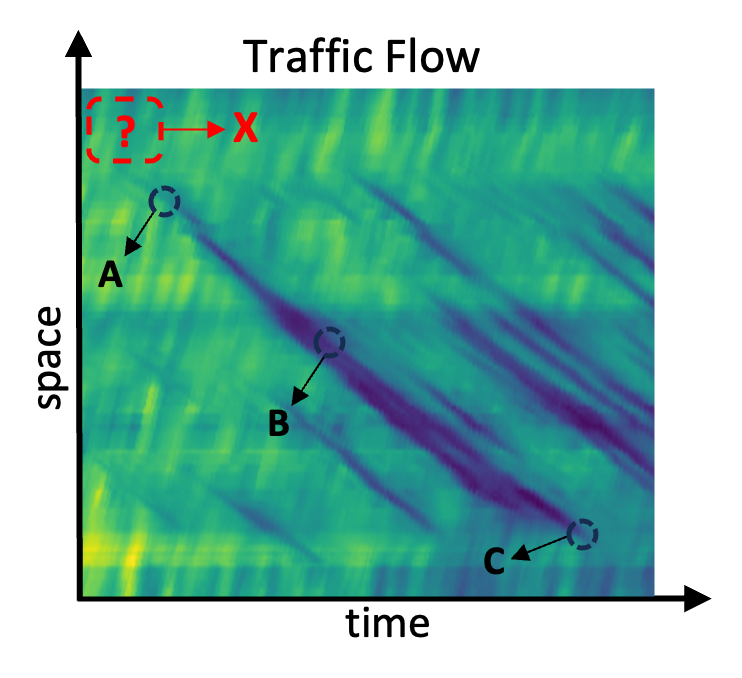}
  \end{center}
  \vspace{-0.7cm}
  \caption{Illustration of traffic breakdowns. A traffic breakdown contains phases of trigger\&formation (A), propagation (B) and dissipation (C). Our goal is to discover the potential traffic breakdown precursors from region X, which is the downstream area antecedent to the breakdown trigger.}
  \vspace{-0.5cm}
  \label{fig:intro}
\end{wrapfigure}

% Understanding the precursors of traffic breakdowns is essential for improving road safety and optimizing traffic flow. 

Traffic breakdowns, characterized by sudden congestion and reduced vehicle speeds, can lead to severe accidents and increased travel times. Identifying the contributing factors enables the development of predictive models to mitigate these events. 

% For example, traffic disturbances caused by driver behavior or external conditions significantly affect stability \cite{jiang2014traffic, yadav2023impact, treiber2007influence, orosz2010traffic, li2006topological}. Connected and automated vehicles (CAVs) offer a promising solution by improving traffic flow stability and reducing fluctuations~\cite{shen2024effects, hung2022impact, zhong2020influence, huang2020scalable, zhou2023autonomous, molnar2024destroying}.

Several methods have been developed to identify and predict traffic breakdowns. Statistical estimators and probabilistic models analyze transitional events, with one approach using statistical estimators to assess breakdown probability by classifying these occurrences~\cite{filipovska2020traffic}. Machine learning techniques, such as artificial neural networks, have also shown promise for modeling abrupt traffic transitions \cite{zhao2023prediction}. However, 
%traditional models often rely on historical data and predefined speed or flow thresholds, which fail to capture the complexity of dynamic, spatially and temporally varying traffic conditions~\cite{grumert2023automated}. 
%Additionally, probabilistic models can produce generalized outcomes that overlook specific situational factors like driver behavior and environmental conditions \cite{filipovska2019prediction, zhao2022traffic, mahendra2019analysis}. 
%While machine learning offers potential, it requires extensive training datasets and may struggle with real-time applications due to computational demands \cite{filipovska2020traffic, barros2015short-term}.
a key limitation of current methods is their inability to systematically link environmental and driver behavior factors with the spatiotemporal dynamics of traffic breakdowns. For instance, while studies highlight precursors such as road geometry or the braking of a lead vehicle in a platoon \cite{mu2021string, tian2017cellular}, input data is often simplified into tabular formats. This simplification discards critical spatial and temporal relationships.
%and overlooks the complex interactions between vehicles and their environment~\cite{saffie2024comparison, feng2023improved, liang2021gis-based}.

Fortunately, recent advancements in spatiotemporal graph neural networks (ST-GNNs) shed light on addressing these gaps by effectively capturing both spatial and temporal dependencies \cite{wang2020traffic, peng2020spatial}. To enhance the interpretability of GNNs, techniques such as Shapley values \cite{winter2002shapley, hart1989shapley} provide robust frameworks for feature attribution in complex models \cite{sundararajan2020many}. Inspired by these techniques, this paper proposes applying ST-GNNs combined with Shapley values to uncover precursors of traffic breakdowns. Specifically, the contributions of this paper are summarized as follows: (1) we extend GNN-based Shapley explanation methods to a spatiotemporal setting, enabling a richer interpretation of traffic dynamics by incorporating both spatial and temporal factors, (2) our proposed ST-GNN Shapley method identifies traffic breakdown precursors by bridging the gap between black-box neural network predictions and human-understandable causes, and (3) we validate our approach through a real-world experiment on I-24, demonstrating its effectiveness and robustness in capturing traffic breakdown precursors. Our findings indicate that the lane adjacent to the merging area is the primary source of traffic fluctuations and congestion.
%and autoencoders are effective for detecting anomalies and identifying unusual patterns leading to breakdowns \cite{zhou2017anomaly, sakurada2014anomaly, an2015variational, chen2018autoencoder}.

% \begin{figure}[h]
% \begin{center}
% \includegraphics[scale=.55]{fig/visual_abstract.png}
% \caption{Illustration of traffic breakdowns. A traffic breakdown contains phases of trigger\&formation (A), propagation (B) and dissipation (C). The precursor of traffic breakdowns will be discovered from region X, which is the downstream area antecedent to the breakdown trigger.} 

% \end{center}
% \end{figure}

% \begin{itemize}
% \item We extend GNN-based Shapley explanation methods to a spatiotemporal setting, allowing for a richer interpretation of traffic dynamics that incorporates both spatial and temporal factors.
% \item Our proposed ST-GNN Shapley method identifies traffic breakdown precursors by bridging the gap between black-box neural network predictions and human-understandable causes.
% \item We demonstrate our approach through a real-world experiment on I-24, demonstrating its effectiveness and robustness in capturing traffic breakdown precursors. We find the lane adjacent to the merging area is the source of traffic fluctuation and jam.
% \end{itemize}

\section{Problem Statement}
Illustrated in Fig.~\ref{fig:intro}, our goal is to identify the potential traffic breakdown precursors in the spatiotemporal domain. To identify the areas and characteristics of the traffic breakdowns, we discretize the road into spatiotemporal cell of fixed spatial and temporal intervals. The road geometry can be represented as an undirected graph $G=(V,E,A)$. Here, $V=\{v_i\}_{i=1}^{N}$ represents the set of nodes with $N=|V|$ indicating the number of nodes. The set of edges is denoted by $E$, and $A \in \mathbb{R}^{N \times N}$ is the adjacency matrix, a square matrix capturing the relationships between the nodes in the graph. In this matrix $A$, each row and column is associated with a node, and the elements $A_{ij}$ specify the existence of edges between the nodes. The features on each node include macroscopic ones like flow, density and speed in each cell, as well as microscopic ones like the trajectory of the vehicles. In this study, we will focus on using the macroscopic features and leave the microscopic ones in the future.

% \subsection{Shapley Value}
% \label{sec:shapley_value}
% \Yanbing{shorten this part}
% The Shapley value is a concept from Game Theory used to fairly distribute the total gains of a game to its players based on their individual contributions, assuming collaboration among all players. It is calculated by averaging each player's marginal contribution across all possible coalitions. This method has been adapted to explain machine learning model predictions for tabular data, treating each feature as a player in a game where the prediction is the reward.

% The characteristic function $val: S \to \mathbb{R}$ measures the marginal contribution of a coalition $S \subseteq \{1, \ldots, F\}$ of features towards the prediction $f(x)$ compared to the average prediction: $val(S) = \mathbb{E}[f(X)|X_S = x_S] - \mathbb{E}[f(X)]$. The impact of feature $j$ is isolated by computing $val(S \cup \{j\}) - val(S)$ and averaging this over all possible ordered coalitions $S$ to obtain its Shapley value:

% \begin{equation}
%     \phi_j(val) = \sum_{S \subseteq \{1, \ldots, F\} \setminus \{j\}} \frac{|S|! (F - |S| - 1)!}{F!} \left( val(S \cup \{j\}) - val(S) \right).
% \end{equation}

% The fairness of the Shapley value is guaranteed by four axioms (efficiency, dummy, symmetry, and additivity), making it the unique solution that satisfies these properties. In practice, the exact computation of Shapley values is infeasible due to the exponential number of possible coalitions, so approximation methods using sampling are employed.

\section{Methodology}

\begin{wrapfigure}{r}{0.5\textwidth}
\vspace{-0.9cm}
  \begin{center}      \includegraphics[width=0.48\textwidth]{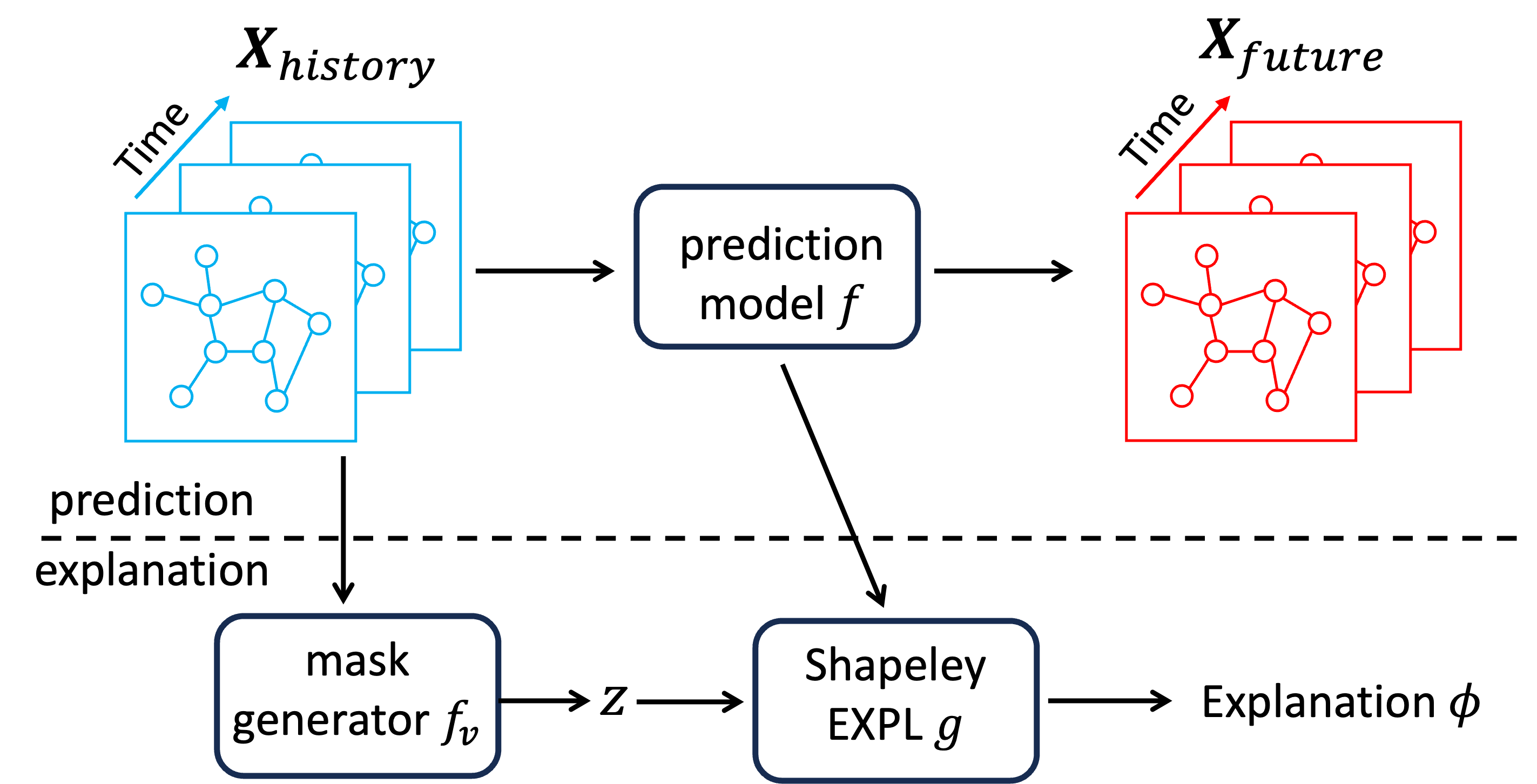}
  \end{center}
  \vspace{-0.5cm}
  \caption{Flowchart of the proposed framework.}
  \vspace{-0.4cm}
  \label{fig:flow}
\end{wrapfigure}
In this section, we introduce the proposed method, which follows a two-step approach. First, we employ a spatiotemporal prediction model to capture traffic dynamics patterns. Second, we develop an explanation model to interpret the predictions and identify precursors of traffic breakdowns. Fig.~\ref{fig:flow} illustrates the structure of the proposed framework. In the upper part, features are input into a spatiotemporal graph neural network for prediction. In the lower part, a mask generator produces a graph mask, which is then used to create ``what-if'' subgraphs by masking node or edge features of the original graph. These subgraphs are fed into a Shapley-based generator, which outputs the explanation $\phi$, identifying the spatiotemporal indices of traffic breakdowns.

\subsection{Mask Generator}
\label{sec:mask_graph_generators}

We begin by developing a mask generator to create discrete masks for spatial and temporal nodes, denoted as \( M_S \in \{0, 1\}^S \) and \( M_T \in \{0, 1\}^T \), respectively. For the instance \( v \) being explained, our goal is to examine the combined effect of a subset of features and its neighbors on the prediction \( f_v(X, A) \). The mask generator identifies the subset to be analyzed. A value of 1 indicates that the variable (node or feature) is included, while 0 indicates it is excluded. First, the mask generator randomly samples from all possible \( 2^{S+T-1} \) pairs of masks \( M_S \) and \( M_T \), representing all potential coalitions of spatiotemporal nodes (excluding \( v \) itself in explanations). Let \( z \) be the random variable for selected nodes, \( z = (M_S \| M_T) \). Then, the ``what-if'' subgraphs \( \tilde{X} \) are generated by masking out the unselected features, \( \tilde{X} = X \odot z \). These subgraphs \( \tilde{X} \) will be used to generate explanations.

\subsection{Spatiotemporal Prediction Model}
There are many choices of the spatiotemporal prediction model $f$. In fact, our proposed Shapley based methods are model-agnostic and can be used to explain ST prediction models. In this paper we adopt the spatiotemporal graph convolutional network (STGCN) as the neural backbone example. The reason of using STGCN is because of its wide application and generalization. 

In STGCN, we first compute the adjacency matrix $A$ of the road topology. The normalized Laplacian matrix is defined as \( L = I - D^{-1/2} A D^{-1/2} \), where \( I \) denotes the identity matrix and \( D \) is the diagonal degree matrix, with each diagonal element \( D_{ii} \) being the sum of the elements in the \( i^{th} \) row of \( A \). To approximate the graph convolution operator \( *_G \), we utilize K-order Chebyshev polynomials, represented by:

\[
g_\theta *_G x = g_\theta(L) x = \sum_{k=0}^{K-1} \theta_k \left(T_k(\tilde{L})\right) x,
\]

\noindent Here, \( \theta \in \mathbb{R}^K \) is the vector of polynomial coefficients, and the scaled Laplacian \( \tilde{L} \) is calculated as \( \tilde{L} = \frac{2}{\lambda_{\max}} L - I \), where \( \lambda_{\max} \) is the largest eigenvalue of \( L \). The Chebyshev polynomials \( T_k(x) \) are defined recursively by \( T_k(x) = 2xT_{k-1}(x) - T_{k-2}(x) \), starting from \( T_0(x) = 1 \) and \( T_1(x) = x \). This K-order Chebyshev polynomial approximation enables each node to be updated based on information from its \( K \) neighboring nodes.

After the graph convolution captures the spatial relationships among neighboring nodes, we apply a standard convolution layer in the temporal dimension to refine the node signals. This temporal convolution layer integrates data across neighboring time slices, updating the node signals accordingly. Finally, a 1×D convolution followed by a nonlinear neural network layer is used to produce the final prediction \( \hat{X}_{(t-\tau+1):t} \). The mean squared error (MSE) is employed as the loss function:

\[
\mathcal{L} = \frac{|| X_{(t-\tau+1):t} - \hat{X}_{(t-\tau+1):t}||^2}{N \tau},
\]

\noindent where \( || \cdot ||^2 \) denotes the \( \mathbb{L}_2 \) norm.

\subsection{Explanation Generator}
\label{sec:explanation_generator}

In this section, we construct a surrogate model \( g \) based on the dataset \( D = \{(z, f(z))\} \) and use it to generate explanations. Formally, an explanation \( \phi \) for \( f \) is selected from a set of potential explanations known as the interpretable domain \( \Omega \). The explanation \( \phi \) is the result of the optimization process:

\begin{equation}
\phi = \arg \min_{g \in \Omega} L_f(g),
\end{equation}

\noindent where the loss function assesses the quality of each explanation. The choice of \( \Omega \) plays a crucial role in determining the type and quality of the explanation produced. In this paper, we broadly define \( \Omega \) as the set of interpretable models, specifically focusing on Weighted Linear Regression (WLR) models. Additionally, we incorporate Shapley values to quantify the contribution of each feature or node in the spatiotemporal graph to the prediction outcome. By computing Shapley values, we ensure a fair attribution of importance among different input variables, providing a principled way to interpret model decisions. For more details on the integration of Shapley values, please refer to \cite{duval2021graphsvx}.

\section{Experiment Results}

\begin{figure}[ht]
    \centering
    % First Row - lane 1 flow
    \begin{subfigure}[b]{0.48\textwidth}
        \centering
        \includegraphics[width=\textwidth]{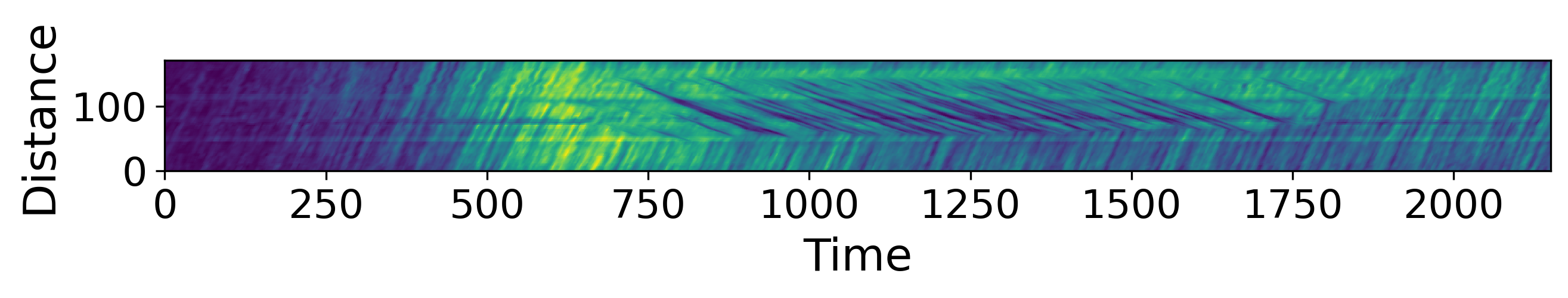}
        \caption{Lane 1 true flow}
    \end{subfigure}
    \hfill
    \begin{subfigure}[b]{0.48\textwidth}
        \centering
        \includegraphics[width=\textwidth]{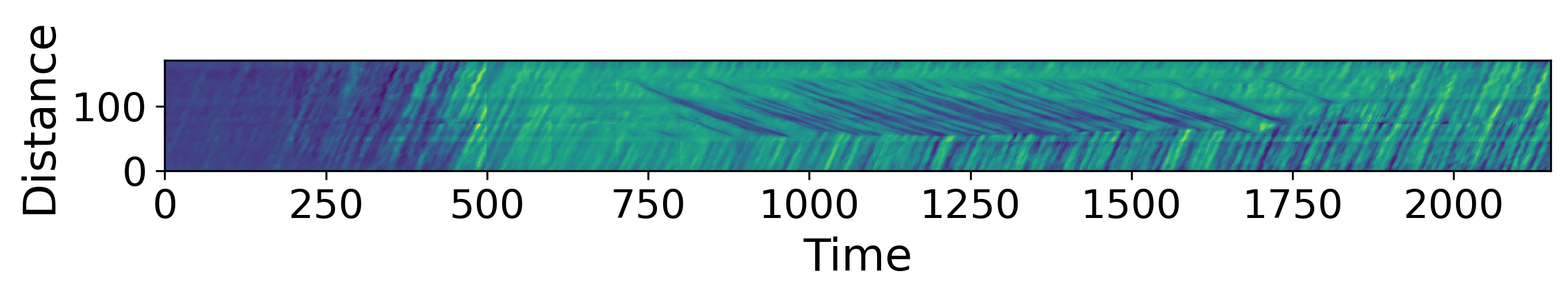}
        \caption{Lane 1 predicted flow}
    \end{subfigure}        
    \caption{Comparison of true and predicted flow and density across four lanes for the I-24 dataset. The distance unit is $0.1$ miles and the time unit is $10$ seconds.}
    \label{fig:true_pred_rds}
\end{figure}

We use macroscopic traffic data (flow, speed, and density) from I-24~\cite{gloudemans202324}, obtained via radar detection systems (RDS) and interpolated using the Adaptive Smoothing Method (ASM)~\cite{treiber2011reconstructing}. The Tennessee Department of Transportation provides the RDS data, collected through a highway surveillance system. Aggregated at 30-second intervals, it includes lane-by-lane occupancy, vehicle counts, and speed. ASM reconstructs smooth, continuous estimates of traffic states by dynamically adjusting smoothing parameters based on local conditions~\cite{ji2024virtual}. The analyzed I-24 segment consists of four main lanes and on/off ramps. Due to data availability, we focus on the four main lanes. The space is divided into $0.1$-mile cells and \( 10 \)-second intervals, forming 684 nodes.

\begin{wrapfigure}{r}{0.35\textwidth}
\vspace{-0.9cm}
\begin{center}
    \centering
    % First Row - True Data
    \begin{subfigure}[b]{0.2\textwidth}
        \centering
        \includegraphics[height=2\textwidth]{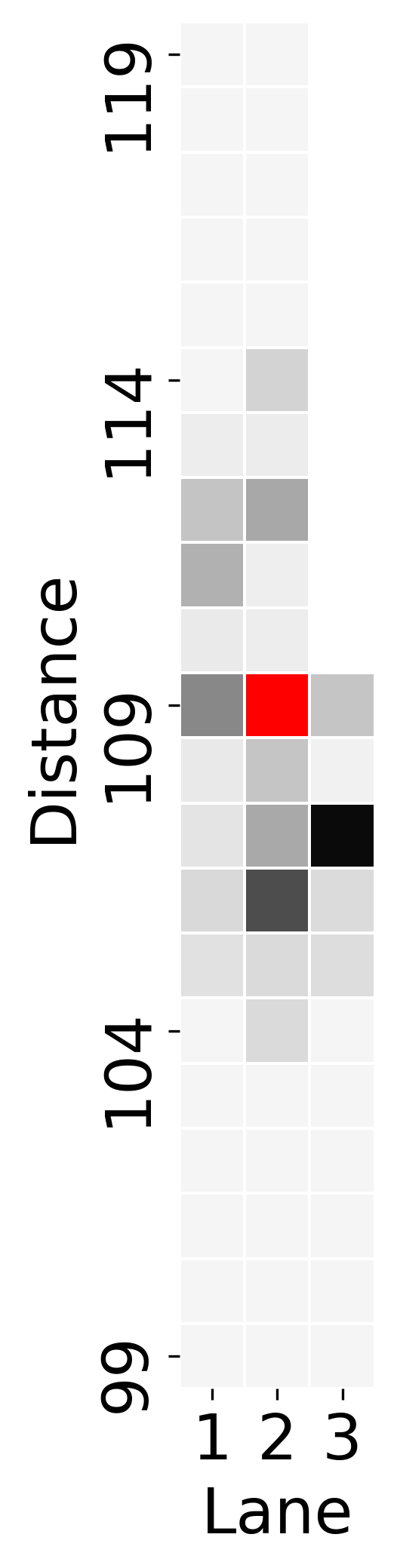}
        %\caption{lane 2}
    \end{subfigure}
    %\hfill
    \hspace{-1.5 cm}
    \begin{subfigure}[b]{0.2\textwidth}
        \centering
        \includegraphics[height=2\textwidth]{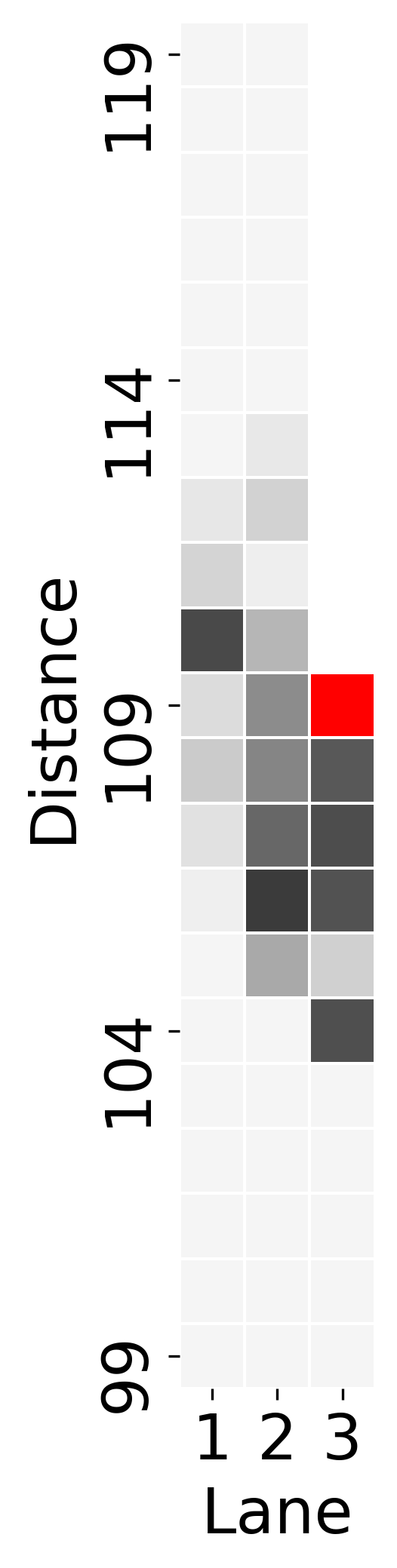}
        %\caption{lane 3}
    \end{subfigure}
    \caption{Spatial distributions of Shapley values among different neighbors for various SAGs. The two subfigures correspond to different investigated nodes, highlighted in red—Lane 2 in (a) and Lane 3 in (b), respectively.}
    \label{fig:spatial_shap}
\end{center}
\vspace{-1.5cm}
\end{wrapfigure}

The prediction results are shown in Fig.~\ref{fig:true_pred_rds}, demonstrating the model's ability to approximate traffic behavior with high accuracy. The predicted flow patterns closely match the ground truth, capturing key variations in traffic dynamics across different lanes. This confirms that the spatiotemporal graph neural network effectively learns the underlying traffic patterns.

The explanation results are presented in Fig.~\ref{fig:spatial_shap}, illustrating the spatial distribution of Shapley values among neighboring nodes, with the red node representing the one under investigation. The results indicate that the most influential neighbors are not necessarily the closest ones, underscoring the impact of road geometry on traffic breakdowns. Specifically, we observe that lane adjacency and merging zones play a significant role in congestion propagation, emphasizing the need to account for spatial dependencies when analyzing traffic breakdown precursors.

\section{Conclusion} 

In this paper, we propose a novel framework combining spatiotemporal graph neural networks (ST-GNNs) with Shapley value-based explanations to identify precursors of traffic breakdowns. Our approach extends existing GNN-based Shapley explanation methods to a spatiotemporal setting, enabling a more interpretable analysis of traffic dynamics. By applying our method to the I-24 dataset, we demonstrate its capability to uncover key contributors to traffic breakdowns, such as lane adjacency effects and abrupt braking patterns. 

Our results highlight that the lane adjacent to the merging area is a major source of traffic fluctuations, often leading to congestion. Additionally, the explanation results reveal that the most influential neighboring nodes are not always the closest ones, underscoring the importance of road topology in traffic breakdown formation. Future work includes extending our model to incorporate microscopic vehicle trajectory data, improving real-time inference capabilities, and exploring broader applications in intelligent traffic management and congestion mitigation strategies.

% \subsection{Traffic-breakdown Identification}
% We use xx for traffic breakdown identification. Its algorithm is shown in xxx. 

%\newpage

% \section{Acknowledgements}
% % The authors would like to thank the Tennessee Department of Transportation for providing the RDS data and Junyi Ji (Vanderbilt University) for providing the validation dataset for the I-24 corridor experiment in this paper. 
% The authors would like to thank the Tennessee Department of Transportation for providing the RDS data and Junyi Ji (Vanderbilt University) for providing the dataset for the I-24 corridor experiment in this paper.
% The submitted manuscript has been created by UChicago Argonne, LLC, Operator of Argonne National Laboratory (“Argonne”). Argonne, a U.S. Department of Energy Office of Science laboratory, is operated under Contract No. DE-AC02-06CH11357. This report and the work described were sponsored by the U.S. Department of Energy (DOE) Vehicle Technologies Office (VTO) under the Energy Efficient Mobility Systems (EEMS) Program, with support from EERE managers Avi Mersky, Erin Boyd and Alexis Zubrow.

% \bibliographystyle{elsarticle-harv}
% \bibliographystyle{abbrv}
\bibliographystyle{unsrt}
\bibliography{survey}

%% first round review opinion

\end{document}